\pdfoutput=1
\documentclass[11pt]{article}

\usepackage{EMNLP2022}

\usepackage{times}
\usepackage{latexsym}

\usepackage[T1]{fontenc}
\usepackage[utf8]{inputenc}

\usepackage{microtype}

\usepackage{inconsolata}
\usepackage{times}
\usepackage{enumitem}
\usepackage{latexsym}
\usepackage{amsfonts}
\usepackage{bm}
\usepackage{bbm}
\usepackage{amsmath}
\usepackage{graphicx}
\usepackage{float}
\usepackage{subfigure} 
\usepackage{verbatim}
\usepackage{multirow}
\usepackage{hyperref}
\usepackage{makecell}
\usepackage[ruled,vlined]{algorithm2e}
\usepackage{makecell}
\usepackage{bm}
\title{Momentum Contrastive Pre-training for Question Answering}

\author{
  Minda Hu$^1$\hspace{0.3cm}Muzhi Li$^1$\hspace{0.3cm}Yasheng Wang$^2$\hspace{0.3cm}Irwin King$^1$\\
  $^1$Dept. of Computer Science \& Engineering, The Chinese University of Hong Kong\\
  $^2$Huawei Noah's Ark Lab\\
  \texttt{\{mindahu21, mzli, king\}@cse.cuhk.edu.hk},\\
  \texttt{wangyasheng@huawei.com}
  }

\newcommand\T{\rule{0pt}{2.6ex}}       % Top strut
\newcommand\B{\rule[-1.0ex]{0pt}{0pt}} % Bottom strut

\begin{document}
\maketitle
\begin{abstract}
Existing pre-training methods for extractive Question Answering (QA) generate cloze-like queries different from natural questions in syntax structure, which could overfit pre-trained models to simple keyword matching. In order to address this problem, we propose a novel \textbf{M}omentum \textbf{C}ontrastive p\textbf{R}e-training f\textbf{O}r que\textbf{S}tion an\textbf{S}wering~(MCROSS) method for extractive QA. Specifically, MCROSS introduces a momentum contrastive learning framework to align the answer probability between cloze-like and natural query-passage sample pairs. Hence, the pre-trained models can better transfer the knowledge learned in cloze-like samples to answering natural questions. Experimental results on three benchmarking QA datasets show that our method achieves noticeable improvement compared with all baselines in both supervised and zero-shot scenarios.
\end{abstract}

\section{Introduction}
The task of extractive Question Answering~(QA), which aims to select an answer span from a passage given a query, is a major focus of NLP research. Currently, deep learning systems~\cite{huang2017fusionnet,devlin2019bert,wu2021recursively} have achieved competitive results with humans on large-scale QA datasets~\cite{rajpurkar2016squad,rajpurkar2018know,yang2018hotpotqa,kwiatkowski2019natural}. However, the collection of high-quality natural QA pairs is still a labor-intensive task, especially for the construction of domain-specific QA systems. To alleviate such data availability restrictions, pre-training methods have been drawing increasing attention~\cite{dhingra2018simple,ram2021few}.

\begin{figure}[!t]
    \setlength{\belowcaptionskip}{-0.5cm}
	\centering
	\includegraphics[width=0.49\textwidth]{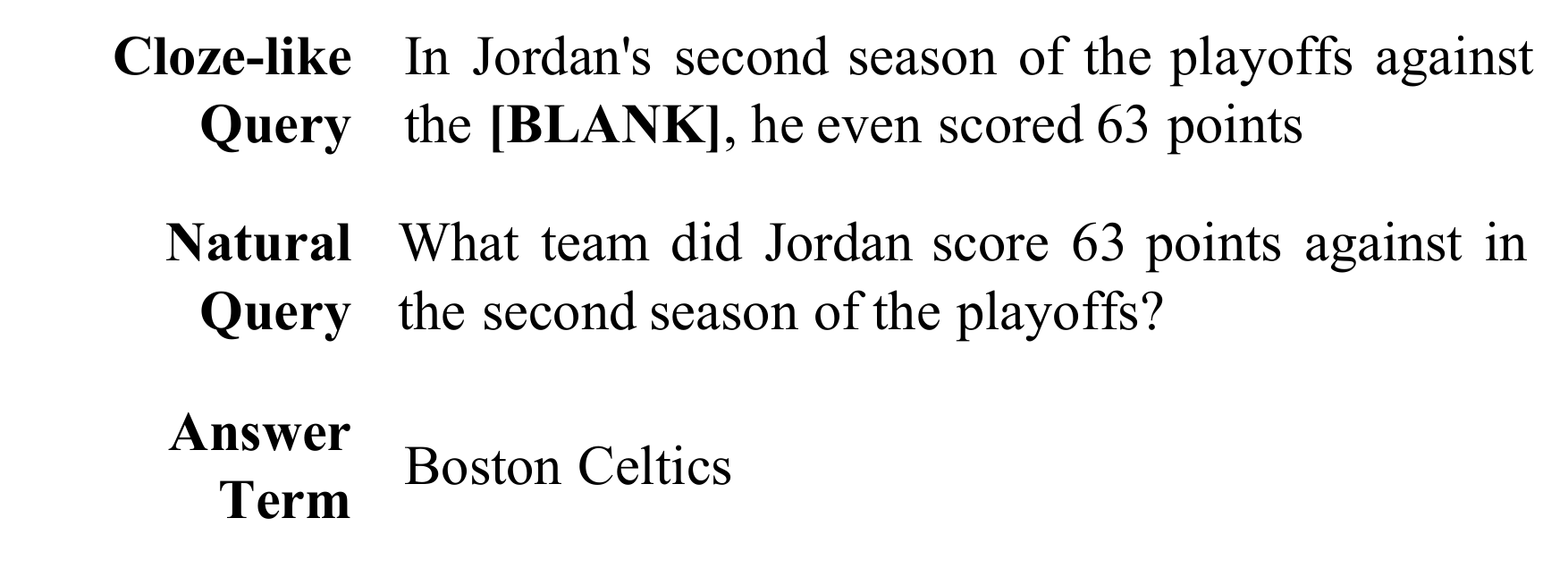}
	\caption{An example of the pre-training queries in SSPT~\cite{glass2019span}. The query is a sentence where a noun phrase or entity~(``Boston Celtics'') is replaced with the special token [BLANK], which differs syntactically from the natural one.}
	\label{fig:noise_data}
\end{figure}

Typically, pre-training methods for extractive QA generate cloze-like query-passage pairs with text-matching techniques. For instance, the Span Selection Pre-Training~(SSPT) method~\cite{glass-etal-2020-span} generates these pairs with the Wikipedia corpus using BM25. Nevertheless, these methods have two major issues. Firstly, the format of cloze-like queries differs much from natural queries asked by humans~(see Fig.~\ref{fig:noise_data}).
%it cannot guarantee that the retrieved passage is relevant to the query. \md{[Consider whether to delete this to fit the story]} In addition, 
%the format of cloze-like queries differs much from %interrogative sentences% 
%natural queries asked by humans. 
In addition, models trained by cloze-like queries overfit in capturing the lexical overlaps between queries and passages, which restricts the capability of high-level semantic reasoning~\cite{Li2020Harvesting,Hu2021Relation}. 
Since answering natural questions requires a comprehensive understanding of queries and passages, models pre-trained with cloze-like queries are not well aligned with downstream QA tasks.
%To tackle this issue, we utilize contrastive learning techniques to circumvent overfitting and better transfer knowledge learned in cloze-like samples into answering natural questions.
 %\wys{Is this a conclusion from previous study? Any reference?}~\mz{Yes, there are conclusion from previous studies, e.g. in the intro part of the 1st reference and related works part of 2nd reference.} 

%The passage is an answer-bearing paragraph retrieved from Wikipedia based on this query. The pre-training task is predicting the answer term from the passage.

%\md{[needs to be changed]}
To solve these issues, we utilize Contrastive Learning~(CL) techniques to align knowledge learned in cloze-like samples to answering natural language questions and circumvent overfitting. 
%Contrastive learning aims to learn representations by contrasting augmentations of different input data, which suits our needs well.
%which has achieved impressive success in the computer vision domain~\cite{He2019moco,Chen2021siamese} 
Specifically, CL aims to learn representations by contrasting augmentations of different input data, which has been successfully applied in various  research fields such as bioinformatics, graph learning, computer vision, and NLP ~\cite{DBLP:journals/bib/HanCCZHCZHCKGL22,DBLP:conf/kdd/ZhangZSKK22,He2019moco,Yang2021xmoco,Pan2021Multilingual}. Nevertheless, existing CL methods for QA focus on query-passage matching~\cite{Yang2021xmoco,Caciularu2022Long} and multilingual embedding alignment~\cite{Pan2021Multilingual}, which have not been tailored to deal with issues of query format inconsistency and overfitting in matching lexical overlaps.
%\mz{I have shortened this paragraph.}%

In this paper, we propose a \emph{Momentum Contrastive pRe-training fOr queStion AnSwering}~(MCROSS) method for extractive QA. Specifically, MCROSS employs a momentum contrastive learning strategy along with the conventional answer prediction task to maximize the consistency of predicted answer distributions between cloze-like and natural query pairs and thus improves the performance of pre-trained models in answering natural language questions.
%to capture the semantic correlation between cloze-like and natural query-passage pairs. \md{[Change citation in this section]} Thus, CROSS adopts a contrastive strategy~\cite{blum1998combining,nigam2000analyzing,han2018co} %
We show the efficacy of our approach on standard English benchmark datasets. On SQuADv1.1~\cite{rajpurkar2016squad}, MCROSS achieves about 2.7/3.5 percentage points gain on F1/Exact Match~(EM) accuracy over BERT, and 1.4/1.8 percentage points improvement on the same metrics compared to SSPT~\cite{glass-etal-2020-span}. MCROSS also consistently outperforms the baseline methods by a large margin on TriviaQA~\cite{joshi2017triviaqa} and NewsQA~\cite{trischler2017newsqa} in supervised and zero-shot scenarios.
\section{Method}
\begin{figure}[!t]
	\centering
	\includegraphics[width=0.49\textwidth]{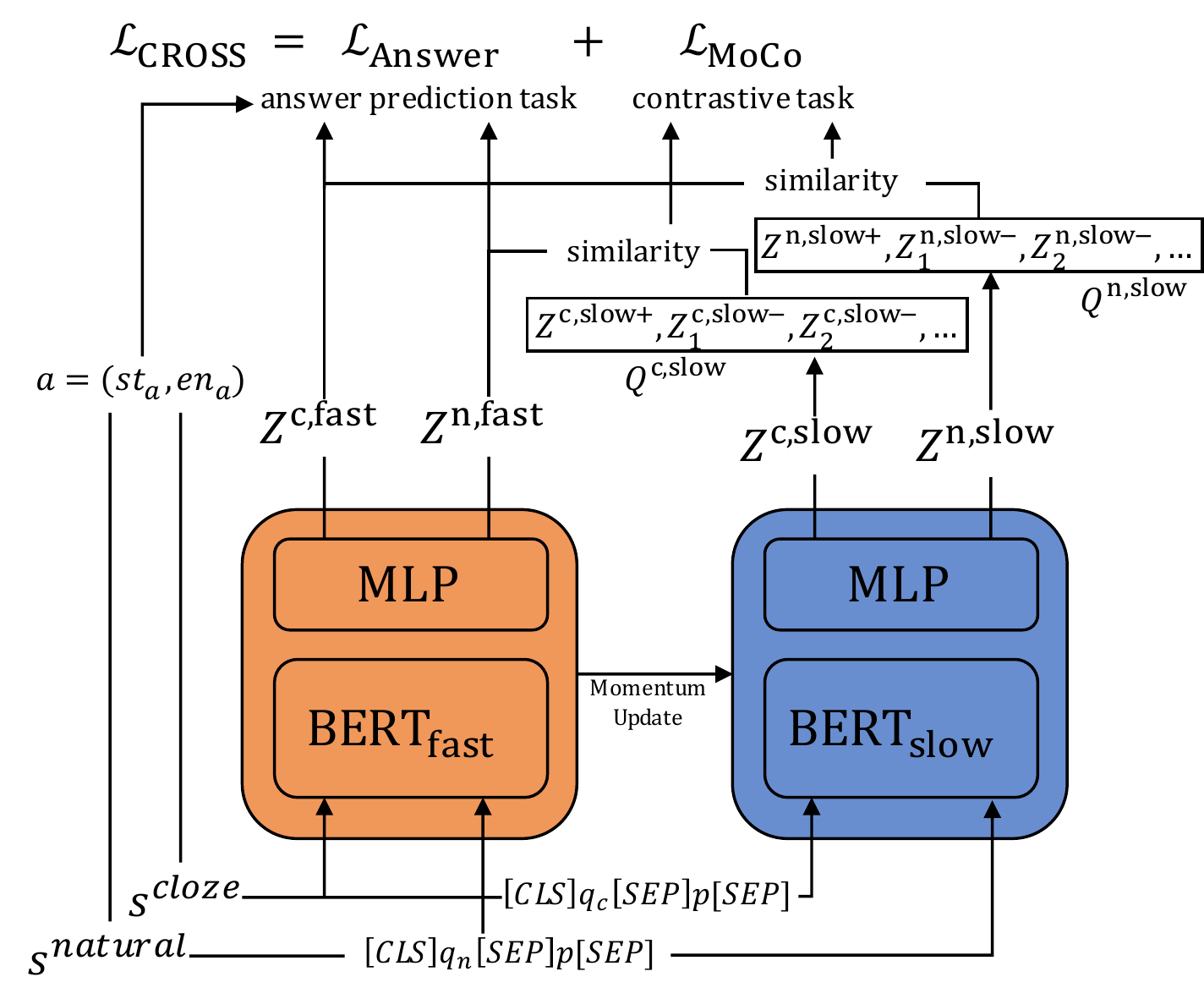}
	\caption{An overview of the proposed MCROSS method.}
	\label{fig:framework}
\end{figure}

The structure of our proposed MCROSS method is shown in Fig.~\ref{fig:framework}. Along with the cloze-like samples $s^{\mathrm{cloze}}=(q_{\mathrm{c}}, p, a)$ generated by SSPT~\cite{glass-etal-2020-span}, we create positive natural samples $s^{\mathrm{natural+}}=(q_{\mathrm{n}}, p, a)$ containing natural queries $q_{\mathrm{n}}$ given passages and answers $(p, a)$ from $s^\mathrm{cloze}$ with T5 answer-aware question generator~\cite{Raffel2020T5} which is fine-tuned with SQuADv1.1 training set. The details of pre-training datasets are illustrated in Appendix~\ref{pretraining_dataset}. Though different in format, the cloze-like queries $q_c$ and positive natural queries $q_n$ are semantically similar since they have the same answers~$a$ in given supporting passages~$p$. Therefore, we expect that the predicted probability distribution of answer span in positive pairs $(s^{\mathrm{cloze}}, s^{\mathrm{natural+}})$ should be closer. On the other hand, we generate negative query pairs $(s^{\mathrm{cloze}}, s^{\mathrm{natural-}})$ with different answers or supporting passages. Correspondingly, the predicted answer span distribution of $s^{\mathrm{natural-}}$ is expected to be far from that of $s^{\mathrm{cloze}}$. MCROSS achieves the goal with two pre-training tasks as follows.

%Generally, CROSS aims to maintain the similarity of output distributions between cloze-like samples $s^{\mathrm{cloze}} = (q_{\mathrm{c}}, p, a)$ and corresponding natural samples $s^{\mathrm{natural}} = (q_{\mathrm{n}}, p, a)$. Although the format are different, the cloze-like query $q_c$ and natural query $q_n$ are semantically similar since they have the same answer $a$ in given passage $p$. Therefore, we expect that the predicted probability distribution of answer span in positive pairs $(s^{\mathrm{cloze}}, s^{\mathrm{natural}})$ with same answer and passage should be more similar than the negative samples $s'_{\mathrm{cloze}}$ and $s'_{\mathrm{natural}}$. CROSS achieves this goal by designing the following two pre-training tasks.

\subsection{Pre-Training Tasks}
% The last sentence of the last paragraph performs similar functionality compared the following sentence
% In this section, we will propose two novel pre-training tasks for CROSS model called contrastive learning and answer term prediction.
% In CROSS, we use two pre-training tasks: answer term prediction task and contrastive learning task.

\begin{description}[leftmargin=0pt,listparindent=\parindent,parsep=0pt]
	\item [Answer Term Prediction Task.] Given a training sample $s=(q,p,a)$, the answer term prediction task aims to predict the token span $a=(st_a, en_a)$ in the query $q$ from the passage $p$\footnote{We treat the first exact match of the token span in the passage as the answer term.}, where $st_a$ and $en_a$ are the start and end positions of the answer term in the passage.
	
    First, we prepare the concatenated token sequence $\mathrm{T} = [CLS]\,q\,[SEP]\,p\,[SEP]$ and leverage BERT~\cite{devlin2019bert} to encode the context embeddings $\bm{H}\in \mathbb{R}^{[\mathrm{len}(q)+\mathrm{len}(p)+3]\times d}$ for each query-passage pair, where $d$ is the dimension of embeddings. Then, the probability distribution $Z = (z^{st_a}, z^{en_a})$ of the start/end indices of answer span can be predicted as:
	\begin{equation}
        %\small
	    \begin{split}
		    z^{st_a}_{i} = p(i=st_a) = \text{softmax}(L^{\mathrm{start}}_{\mathrm{span}}(\bm{H}_i)),\\
		    z^{en_a}_{i} = p(i=en_a) = \text{softmax}(L^{\mathrm{end}}_{\mathrm{span}}(\bm{H}_i)),
	    \end{split}
	    \label{eq:answer_span_predict}
    \end{equation}
    where $L^{\mathrm{start}}_{\mathrm{span}}$ and $L^{\mathrm{end}}_{\mathrm{span}}$ are linear answer span prediction layers. The loss function of the answer term prediction task is defined below:
    \begin{equation}
        %\small
	    \begin{aligned}
	        \mathcal{L}_{\text{Answer}}&(Z) = -\frac{1}{2}\{ \sum_{1\leq i \leq N} \mathbbm{1}(i=st_a)\log z^{st_a}_{i} \\
	        & + \sum_{1\leq i \leq N} \mathbbm{1}(i=en_a)\log z^{en_a}_{i} \},
	    \end{aligned}
	    \label{eq:loss_answer}
    \end{equation}
    where $\mathbbm{1}(\text{condition})$ is the indicator function that returns 1 if the condition is satisfied and 0 otherwise. $N$ is the length of the sequence $\mathrm{T}$.
	\item [Contrastive Learning Task.] 
	This task utilizes a contrastive loss function to guide the answer span distributions $(Z^{\mathrm{cloze}}, Z^{\mathrm{natural+}})$ of positive sample pairs $(s^{\mathrm{cloze}}, s^{\mathrm{natural+}})$ with same $(p, a)$ to be closer, while keeping $Z^{\mathrm{cloze}}$ to be dissimilar from $Z^{\mathrm{natural-}}$ of $s^{\mathrm{natural-}}$ with different passages $p^-$ or answers $a^-$.
	%This task aims to guide the answer span distributions $z^{\mathrm{cloze}}$ of $s^{\mathrm{cloze}}$  to be closely resemble to the counterparts $z^{\mathrm{natural}}$ of $s^{\mathrm{natural}}$ with same $(p, a)$ while encouraging $z_{\mathrm{cloze}}$ to be dissimilar from $z'_{\mathrm{natural}}$ of $s'_{\mathrm{natural}}$ with $(p', a')$ different from $(p, a)$ through a contrastive loss. %
	
	%\mz{It is better to place the following sentence into another position, which will make more sense.}
	%Using \textbf{Mo}mentum \textbf{Co}nstrastive learning~(MoCo), we expand the pool of negative samples which enable our model to learn better representations.%
	
	%\wys{related works should be added?}
	
	Inspired by the \textbf{Mo}mentum \textbf{Co}nstrastive learning~(MoCo)~\cite{He2019moco} strategy, we maintain a large pool of consistent negative samples from several previous batch iterations to preserve more information on negative samples and obtain better pre-training objectives. Following MoCo, we employ a dual-encoder architecture $(\mathrm{BERT}_{\mathrm{fast}}, \mathrm{BERT}_{\mathrm{slow}})$ and maintain FIFO queues $(Q^{\mathrm{slow}}_{st_a}, Q^{\mathrm{slow}}_{en_a})$ containing a large set of answer span distributions $(z^{\mathrm{slow-}}_{st_a}, z^{\mathrm{slow-}}_{en_a})$ of negative samples predicted by $\mathrm{BERT}_{\mathrm{slow}}$. To maintain queue consistency, for each batch iteration, parameters $\theta_{\mathrm{slow}}$ of $\mathrm{BERT}_{\mathrm{slow}}$ are only updated by the exponential moving average of $\theta_{{\mathrm{fast}}}$ of $\mathrm{BERT}_{\mathrm{fast}}$. Here we have
	\begin{equation}
        %\small
	    \begin{split}
		    \theta_{{\mathrm{slow}}}=m \theta_{{\mathrm{slow}}}+(1-m) \theta_{{\mathrm{fast}}},
	    \end{split}
	    \label{eq:momentum_update}
    \end{equation}
    where $m$ is the momentum coefficient. We regulate the output of the answer-span classification layer with InfoNCE loss to bring positive pairs $(s, s^+)$ closer to each other and push negative ones apart:
	\begin{equation}
        %\small
	    \begin{aligned}
		    &\mathcal{L}_{\mathrm{MoCo}}(z^\mathrm{fast}, z^{\mathrm{slow+}}, Q) =
		    \\ & -\log \frac{\exp \left(\mathcal{D}(z^{\mathrm{fast}}, z^{\mathrm{slow+}}) / \tau\right)}{\sum_{z^{\mathrm{slow-}} \in Q} \exp \left(\mathcal{D}(z^{\mathrm{fast}}, z^{\mathrm{slow-}}) / \tau\right)},
	    \end{aligned}
	    \label{eq:infonce_loss}
    \end{equation}
    where $z^{\mathrm{fast}}$ and $z^{\mathrm{slow}+}$ represent the start or end distributions of positive pairs, $\tau$ is
    the softmax temperature, and $Q$ denotes queues containing distributions $z^\mathrm{slow-}$ of negative samples $s^-$ which have $(p^-, a^-)$ being different from $s^+$. $\mathcal{D}$ is the similarity function measuring distances between answer span distributions. 
    Cosine similarity is utilized in the original MoCo method~\cite{He2019moco}. However, we argue that KL-Divergence is more suitable for the measurement of differences between two probability distributions, which is evaluated and validated in Appendix~\ref{design_choice}. %\wys{?}
\end{description}

%\subsection{Data Augmentation}
\subsection{Variants of the MCROSS Method}\label{sec:cross_variants}
There are two variants of the MCROSS method with different loss functions, the first is involved with the unilateral loss function, and the second one with the bilateral loss.
\begin{description}[leftmargin=0pt,listparindent=\parindent,parsep=0pt]
%\item [CROSS~(w/o $\boldsymbol{\mathcal{L}}_{\boldsymbol{\mathrm{MoCo}}}$)] 
\item [MCROSS~(UNI).]
    In addition to the original SSPT training loss $\mathcal{L}_{\text{Answer}}(Z^\mathrm{c})$, we add $\mathcal{L}_{\text{MoCo}}$ to maximize the consistency of prediction between $s^{\mathrm{cloze}}$ and $s^{\mathrm{natural}}$ with little overhead in the unilateral version of MCROSS. Here, we have the unilateral MoCo loss
    \begin{equation}
        %\small
	    \begin{aligned}
	        &\mathcal{L}_{\text{UNI}}(s^{\mathrm{cloze}}, s^{\mathrm{natural}})  =  \mathcal{L}_{\text{Answer}}(Z^\mathrm{c}) \\ & + \lambda_{\mathrm{MoCo}} / 2 * [\mathcal{L}_{\text{MoCo}}(z^{st_a,\mathrm{c}}, z^{st_a,\mathrm{n}}, Q^{\mathrm{n,slow}}_{st_a}) \\ & + \mathcal{L}_{\text{MoCo}}(z^{en_a,\mathrm{c}}, z^{en_a,\mathrm{n}}, Q^{\mathrm{n,slow}}_{en_a})],
	    \end{aligned}
	    \label{eq:uni_loss_context}
    \end{equation}
    where $Z^\mathrm{n}=(z^{st_a,\mathrm{n}}, z^{en_a,\mathrm{n}})$ are the predicted distributions of start/end indices for $s^\mathrm{natural}$, and $Z^\mathrm{c}=(z^{st_a,\mathrm{c}}, z^{en_a,\mathrm{c}})$ are for $s^\mathrm{cloze}$. $(Z^\mathrm{c},Z^\mathrm{n})$ are predicted by $\mathrm{BERT}_{\mathrm{fast}}$ and $,\mathrm{BERT}_{\mathrm{slow}}$, respectively. $Z^\mathrm{n}$ are stored in $(Q^{\mathrm{n,slow}}_{st_a}, Q^{\mathrm{n,slow}}_{en_a})$ and will be reused in later training steps as negative examples. $\lambda_{\mathrm{MoCo}}$ denotes the ratio of MoCo loss.
%\mz{I think the following description is better}
\item [MCROSS~(BI).]
    In MCROSS~(UNI) method, only cloze-like samples $s^{\mathrm{cloze}}$ are considered in the calculation of answer term prediction loss $\mathcal{L}_{\text{Answer}}$, which makes the fast encoder $\mathrm{BERT}_{\mathrm{fast}}$ to be less sensitive to natural samples $s^{\mathrm{natural}}$. Therefore, it will be difficult to properly maintain the consistency between the probability distributions generated from the two types of samples.
    % In CROSS~(UNI) method, only cloze-like queries $s^{\mathrm{cloze}}$ are utilized in the answer term prediction task, the corresponding natural queries $s^{\mathrm{natural}}$ will not be considered in the optimization of answer-span classification layer in $\mathrm{BERT}_q$. 
    % since $\mathrm{BERT}_q$ only receives cloze-like queries $s^{\mathrm{cloze}}$ while natural queries $s^{\mathrm{natural}}$ will not be directly utilized in answer span prediction task. 
    % In CROSS~(UNI) method, as the $\mathrm{BERT}_q$ only sees $s^{\mathrm{cloze}}$ directly and $s^{\mathrm{natural}}$ encoded by slow encoder $\mathrm{BERT}_k$ does not have straightforward influence on the gradients, the training becomes insensitive to the passage representations and is unable to properly optimize the consistency. 
    To tackle this issue, following xMoCo~\cite{Yang2021xmoco}, we leverage two $\mathcal{L}_{\mathrm{UNI}}$ losses and jointly optimize the alignment of $(s^{\mathrm{cloze}}, s^{\mathrm{natural}})$ and $(s^{\mathrm{natural}}, s^{\mathrm{cloze}})$ in the following bilateral loss.
    \begin{equation}
        %\small
	    \begin{aligned}
	        \mathcal{L}_{\text{BI}}  = & \ \mathcal{L}_{\text{UNI}}(s^{\mathrm{cloze}}, s^{\mathrm{natural}})
	        \\ & + \mathcal{L}_{\text{UNI}}(s^{\mathrm{natural}}, s^{\mathrm{cloze}}),
	    \end{aligned}
	    \label{eq:bi_loss_context}
    \end{equation}
    where distributions of $s^{\mathrm{natural}}$ are obtained from $\mathrm{BERT}_{\mathrm{slow}}$ and stored in queue $(Q^{\mathrm{n,slow}}_{st_a}, Q^{\mathrm{n,slow}}_{en_a})$ for $\mathcal{L}_{\text{UNI}}(s^{\mathrm{cloze}}, s^{\mathrm{natural}})$. The same practices are applied to $s^{\mathrm{cloze}}$ for $\mathcal{L}_{\text{UNI}}(s^{\mathrm{natural}}, s^{\mathrm{cloze}})$.
    %, and $Z^\mathrm{c}$ of $s^\mathrm{cloze}$ are acquired from $\mathrm{BERT}_{\mathrm{slow}}$.
    
    We also evaluate MCROSS~(w/o $\mathcal{L}_{\mathrm{MoCo}}$), which utilizes multi-task answer prediction loss to predict similar span $a$ given $(q_c, p)$ and $(q_n, p)$ with $\mathrm{BERT}_{\mathrm{fast}}$:
    \begin{equation}
        %\small
	    \begin{aligned}
	        \mathcal{L}_{\text{MCROSS}} = &\  \mathcal{L}_{\text{Answer}}(Z^{\mathrm{n}})+\mathcal{L}_{\text{Answer}}(Z^{\mathrm{c}}).
	    \end{aligned}
	    \label{eq:cross_without_moco}
    \end{equation}
    %where $Z^{\mathrm{n}}$ is the distribution of answer term for $s^\mathrm{natural}$ and $Z^{\mathrm{c}}$ is for $s^\mathrm{cloze}$.
\end{description} 
%Since the test set of SQuAD is not available, we split the original training set (9:1) into training and validation set, and use the original validation set as the test set.

\begin{table}[!t]
\small
\centering
\setlength{\abovecaptionskip}{0.1cm}
\setlength{\belowcaptionskip}{-0.5cm}
\begin{tabular}{ccccc}
\hline
\thead{\textbf{Method}} & \thead{\textbf{F1}} & \thead{\textbf{EM}} &
\thead{\textbf{0-shot}\\\textbf{F1}} &
\thead{\textbf{0-shot}\\\textbf{EM}}\T\B\\
\hline
BERT & 87.43 & 79.50 & 8.00 & 0.06 \T\B\\
SSPT & 88.75 & 81.25 & 22.88 & 15.42\B\\
SSPT$^\dag$ & 89.25 & 82.13 & 64.41 & 44.44\B\\
\hline
MCROSS(UNI) & 88.95 & 81.60 & 24.78 & 16.58\T\B\\
%SSPT(64) & 88.64 & 80.96 & 20.30 & 12.13  \\
%SSPT(64)$^\dag$ & 89.77 & 82.36 & 63.07 & 41.84 \\
MCROSS(BI) & \textbf{90.11} & \textbf{83.03} & \textbf{65.68} & \textbf{45.40} \B\\
w/o $\mathcal{L}_{\mathrm{MoCo}}$ & 89.92 & 82.96 & 65.20 & 44.82 \B\\
\hline
\end{tabular}
\caption{\label{squad-table}Results on SQuADv1.1 dataset.~(in domain)} 
\end{table}

\section{Experiments} 
We use the following three English span-extraction QA datasets to evaluate pre-trained models on F1/EM metrics in both supervised and zero-shot scenarios. Specifically, TriviaQA~\cite{joshi2017triviaqa} and NewsQA~\cite{trischler2017newsqa} are out-of-domain datasets from MRQA 2019 Shared Task~\cite{fisch2019mrqa}. Since  SQuADv1.1~\cite{rajpurkar2016squad} dataset is utilized to fine-tune the T5 question generator, it is also included to examine the in-domain QA performance.

The experimental settings, implementation details, and complexity analysis are presented in Appendix \ref{experimental_setting}, \ref{implementation_detail}, and \ref{space_time_complexity}.
\subsection{Baselines}
We compare the MCROSS with the following four baselines:
\begin{itemize}[leftmargin=*,parsep=0pt,itemsep=0pt,topsep=0pt]
\item BERT: The 12-layer BERT model released by~\cite{devlin2019bert}.
\item MRQA-BERT: The official multi-task baseline BERT-base model from MRQA\footnote{\url{https://github.com/mrqa/MRQA-Shared-Task-2019/tree/master/baseline}}.
\item SSPT: The span selection pre-training method proposed by~\cite{glass2019span}. This method trains models with cloze-like samples $s^{\mathrm{cloze}}$ using answer term prediction loss $\mathcal{L}_{\text{Answer}}(Z^\mathrm{c})$ in Eq.~(\ref{eq:cross_without_moco}).
\item SSPT$^\dag$: SSPT method trained with only natural samples $s^{\mathrm{natural}}$ using answer term prediction loss $\mathcal{L}_{\text{Answer}}(Z^\mathrm{n})$ in Eq.~(\ref{eq:cross_without_moco}). 
\end{itemize}
%SSPT and SSPT$^\dag$ are identical to CROSS(UNI) in the number of training steps and samples. We also include SSPT(64) and SSPT(64)$^\dag$ with batch size of 64 per iteration, which is equivalent to CROSS(BI) and CROSS(w/o $\mathcal{L}_{\text{MoCo}}$).

\subsection{Experiment Results}
\begin{description}[leftmargin=0pt,listparindent=\parindent,parsep=0pt]
\item[SQuADv1.1.]
Table~\ref{squad-table} shows the performance of all models on SQuADv1.1. Compared with the BERT baseline without extended pre-training, three variants of the proposed MCROSS method increase the F1/EM metrics by at least 1.5/2.1 percentage points. In addition, MCROSS(w/o $\mathcal{L}_{\mathrm{MoCo}}$) achieves discernible boosts among all metrics compared to baseline SSPT. It proves that the combination of cloze-like and natural samples in the QA task can endow the pre-trained model with a better understanding on supporting passages. 
% It proves that $s^{\mathrm{cloze}}$ and $s^{\mathrm{natural}}$ datasets are complementary to each other, and the answer prediction task can endow the pre-trained model with a better understanding of QA task. 
Moreover, the gap between MCROSS(BI) and MCROSS(UNI) indicates that models trained only with cloze-like samples will be insensitive to natural questions. It should be also noticed that there exists a performance gap between SSPT$^\dag$ and MCROSS(UNI) among all metrics. This is attributed to the fact that the natural samples used in SSPT$^\dag$ are generated by the T5 question generator~\cite{Raffel2020T5}. Since the T5 generator is fine-tuned on the training set of SQuADv1.1, it is well fitted in the domain of SQuAD. Compared to MCROSS(UNI) using cloze-like samples, SSPT$^\dag$ is trained with the natural samples containing more domain knowledge of the dataset, thus surpassing MCROSS(UNI). 
% , we speculate the gap is caused by the knowledge being  distilled into SSPT$^\dag$ from the T5-base question generator. With $s^{\mathrm{natural}}$ generated by the T5-base, SSPT$^\dag$ is well fitted into the domain of SQuAD. Therefore, it surpasses CROSS(UNI) in which $s_{\mathrm{natural}}$ affects the training indirectly through $\mathcal{L}_{\mathrm{MoCo}}$.

\item[NewsQA and TriviaQA.]
Table~\ref{newsqa-table} and Table~\ref{triviaqa-table} show the results on NewsQA and TriviaQA dataset from MRQA 2019 Shared Task. It is noticeable that MCROSS(BI) performs the best among all methods on F1 metrics. Compared with the state-of-the-art baseline SSPT, it achieves an improvement of 2.0 F1 score and 1.6 EM accuracy on NewsQA. On the TriviaQA dataset, MCROSS(BI) also surpasses SSPT by 2.8 percentage points on F1 and EM accuracy. Furthermore, the improvement on TriviaQA of MCROSS(BI) over MCROSS(w/o $\mathcal{L}_{\text{MoCo}}$) demonstrates the effectiveness of $\mathcal{L}_{\text{MoCo}}$ in the out-of-domain setting. In both datasets, the great improvement on zero-shot F1/EM of SSPT$^\dag$ over SSPT exhibits that SSPT$^\dag$ has gained the domain knowledge in understanding natural questions from SQuAD. In comparison with SSPT$^\dag$, MCROSS(BI) significantly boosts zero-shot QA performance by 5.6/6.2 percentage points of F1/EM accuracy.

In contrast to the SQuAD dataset with in-domain settings, MCROSS(UNI) has gained a noticeable advantage over SSPT$^\dag$ on both NewsQA and TriviaQA datasets, hinting that SSPT$^\dag$ is overfitted to the natural samples of SQuAD with extra domain knowledge. Interestingly, MCROSS(BI) performs worse in zero-shot scenarios on NewsQA dataset than MCROSS(w/o $\mathcal{L}_{\text{MoCo}}$). This may be due to the fact that NewsQA has 32.7\% samples that can be easily answered by simple text-matching~\cite{trischler2017newsqa}, into which MCROSS(w/o $\mathcal{L}_{\mathrm{MoCo}}$) is overfitted to capture lexical overlaps.
\end{description}

\begin{table}[!t]
\small
\centering
\setlength{\abovecaptionskip}{0.1cm}
\setlength{\belowcaptionskip}{-0.3cm}
\begin{tabular}{ccccc}
\hline
\thead{\textbf{Method}} & \thead{\textbf{F1}} & \thead{\textbf{EM}} &
\thead{\textbf{0-shot}\\\textbf{F1}} &
\thead{\textbf{0-shot}\\\textbf{EM}}\T\B\\
\hline
BERT & 64.28 & 49.17 & 3.55 & 0.02\T\B\\
SSPT & 65.86 & 50.62 & 9.58 & 6.05\B\\
SSPT$^\dag$ & 65.60 & 50.52 & 34.03 & 15.67\B\\
MRQA-BERT & 66.80 & 50.80 & N/A & N/A\B\\
\hline
MCROSS(UNI) & 66.45 & 51.33 & 10.90 & 6.55\T\B\\
%SSPT(64) & 66.29 & 51.35 & 7.43 & 4.15  \\
%SSPT(64)$^\dag$ & 67.06 & \textbf{52.21} & 32.73 & 13.11  \\
MCROSS(BI) & \textbf{67.90} & \textbf{52.18} & 34.82 & 14.98\B\\
w/o $\mathcal{L}_{\mathrm{MoCo}}$ & 66.95 & 52.14 & \textbf{35.68} & \textbf{16.50}\B\\
%\md{New w/o $\mathcal{L}_{\mathrm{MoCo}}$} & 66.99 & 51.52 & 34.96 & 15.19 \\
\hline
\end{tabular}
\caption{\label{newsqa-table}Results on NewsQA dataset.~(out of domain)}
\end{table}

\begin{table}[!t]
\small
\centering
\setlength{\abovecaptionskip}{0.1cm}
\setlength{\belowcaptionskip}{-0.3cm}
\begin{tabular}{ccccc}
\hline
\thead{\textbf{Method}} & \thead{\textbf{F1}} & \thead{\textbf{EM}} &
\thead{\textbf{0-shot}\\\textbf{F1}} &
\thead{\textbf{0-shot}\\\textbf{EM}}\T\B\\
\hline
BERT & 62.86 & 57.39 & 3.25 & 0.04 \T\B\\
SSPT & 70.93 & 65.27 & 26.89 & 22.47\B\\
SSPT$^\dag$ & 69.09 & 63.24 & 42.01 & 33.99\B\\
MRQA-BERT & 71.60 & 65.60 & N/A & N/A\B\\
\hline
%SSPT(64) & 71.29 & 65.79 & 19.25 & 15.14  \\
%SSPT(64)$^\dag$ & 70.97 & 65.27 & 42.16 & 33.29  \\
MCROSS(UNI) & 72.06 & 66.41 & 28.28 & 23.48\T\B\\
MCROSS(BI) & \textbf{73.77} & \textbf{68.05} & \textbf{47.79} & \textbf{39.53}\B\\
w/o $\mathcal{L}_{\mathrm{MoCo}}$ & 73.30 & 67.72 & 46.63 & 38.54\B\\
%\md{New w/o $\mathcal{L}_{\mathrm{MoCo}}$} & 73.21 & 67.35 & 46.76 & 38.46 \\
\hline
\end{tabular}
\caption{\label{triviaqa-table}Results on TriviaQA dataset.~(out of domain)}
\end{table}

\section{Conclusion}
This paper presents a novel pre-training method MCROSS for extractive QA which contains two tasks: 1) contrastive learning and 2) answer term prediction. Specifically, MCROSS adapts MoCo frameworks to maintain consistency in answering cloze-like and natural questions, enabling pre-trained models to have a more comprehensive understanding of supporting passages. The empirical experiments on three public datasets demonstrate that our approach can obtain noticeable improvements in extractive QA tasks in supervised and zero-shot scenarios.
\section{Limitations}
Although MCROSS can already obtain satisfactory QA performance, due to limited time and computational resources, we only use 5 million cloze-like samples for pre-training, which is one-twentieth of the scale of original SSPT experiments. 
% Furthermore, the question generator is fine-tuned with just one dataset~(SQuADv1.1), limiting the diversity of generated questions. More diversified natural questions and a larger experiment scale could lead to more performance boost in comparison with our current results. We will explore these promising research directions in future works. In addition, experiments should be done with more datasets in multiple languages to further prove the robustness of the CROSS method. 
\section{Acknowledgement}
The work described here was partially supported by grants from the RGC General Research Funding Scheme (GRF) 14222922 (CUHK 2151185).
\bibliographystyle{acl_natbib}
\bibliography{emnlp2022,acl2021}

\begin{thebibliography}{23}
\expandafter\ifx\csname natexlab\endcsname\relax\def\natexlab#1{#1}\fi

\bibitem[{Caciularu et~al.(2022)Caciularu, Dagan, Goldberger, and
  Cohan}]{Caciularu2022Long}
Avi Caciularu, Ido Dagan, Jacob Goldberger, and Arman Cohan. 2022.
\newblock \href {https://doi.org/10.18653/v1/2022.naacl-main.207} {Long context
  question answering via supervised contrastive learning}.
\newblock In \emph{Proceedings of the 2022 Conference of the North American
  Chapter of the Association for Computational Linguistics: Human Language
  Technologies}, pages 2872--2879, Seattle, United States. Association for
  Computational Linguistics.

\bibitem[{Devlin et~al.(2019)Devlin, Chang, Lee, and
  Toutanova}]{devlin2019bert}
Jacob Devlin, Ming-Wei Chang, Kenton Lee, and Kristina Toutanova. 2019.
\newblock Bert: Pre-training of deep bidirectional transformers for language
  understanding.
\newblock In \emph{NAACL-HLT (1)}.

\bibitem[{Dhingra et~al.(2018)Dhingra, Danish, and
  Rajagopal}]{dhingra2018simple}
Bhuwan Dhingra, Danish Danish, and Dheeraj Rajagopal. 2018.
\newblock Simple and effective semi-supervised question answering.
\newblock In \emph{Proceedings of the 2018 Conference of the North American
  Chapter of the Association for Computational Linguistics: Human Language
  Technologies, Volume 2 (Short Papers)}, pages 582--587.

\bibitem[{Fisch et~al.(2019)Fisch, Talmor, Jia, Seo, Choi, and
  Chen}]{fisch2019mrqa}
Adam Fisch, Alon Talmor, Robin Jia, Minjoon Seo, Eunsol Choi, and Danqi Chen.
  2019.
\newblock {MRQA} 2019 shared task: Evaluating generalization in reading
  comprehension.
\newblock In \emph{Proceedings of 2nd Machine Reading for Reading Comprehension
  (MRQA) Workshop at EMNLP}.

\bibitem[{Glass et~al.(2020{\natexlab{a}})Glass, Gliozzo, Chakravarti,
  Ferritto, Pan, Bhargav, Garg, and Sil}]{glass-etal-2020-span}
Michael Glass, Alfio Gliozzo, Rishav Chakravarti, Anthony Ferritto, Lin Pan,
  G~P~Shrivatsa Bhargav, Dinesh Garg, and Avi Sil. 2020{\natexlab{a}}.
\newblock \href {https://doi.org/10.18653/v1/2020.acl-main.247} {Span selection
  pre-training for question answering}.
\newblock In \emph{Proceedings of the 58th Annual Meeting of the Association
  for Computational Linguistics}, pages 2773--2782, Online. Association for
  Computational Linguistics.

\bibitem[{Glass et~al.(2020{\natexlab{b}})Glass, Gliozzo, Chakravarti,
  Ferritto, Pan, Bhargav, Garg, and Sil}]{glass2019span}
Michael Glass, Alfio Gliozzo, Rishav Chakravarti, Anthony Ferritto, Lin Pan,
  G~P~Shrivatsa Bhargav, Dinesh Garg, and Avi Sil. 2020{\natexlab{b}}.
\newblock \href {https://doi.org/10.18653/v1/2020.acl-main.247} {Span selection
  pre-training for question answering}.
\newblock In \emph{Proceedings of the 58th Annual Meeting of the Association
  for Computational Linguistics}, pages 2773--2782, Online. Association for
  Computational Linguistics.

\bibitem[{Han et~al.(2022)Han, Cheng, Chen, Zhong, Hu, Chen, Zong, Hong, Chan,
  King, Gao, and Li}]{DBLP:journals/bib/HanCCZHCZHCKGL22}
Wenkai Han, Yuqi Cheng, Jiayang Chen, Huawen Zhong, Zhihang Hu, Siyuan Chen,
  Licheng Zong, Liang Hong, Ting{-}Fung Chan, Irwin King, Xin Gao, and Yu~Li.
  2022.
\newblock \href {https://doi.org/10.1093/bib/bbac377} {Self-supervised
  contrastive learning for integrative single cell rna-seq data analysis}.
\newblock \emph{Briefings Bioinform.}, 23(5).

\bibitem[{He et~al.(2020)He, Fan, Wu, Xie, and Girshick}]{He2019moco}
Kaiming He, Haoqi Fan, Yuxin Wu, Saining Xie, and Ross Girshick. 2020.
\newblock \href {https://doi.org/10.1109/CVPR42600.2020.00975} {Momentum
  contrast for unsupervised visual representation learning}.
\newblock In \emph{2020 IEEE/CVF Conference on Computer Vision and Pattern
  Recognition (CVPR)}, pages 9726--9735.

\bibitem[{Hu et~al.(2021)Hu, Sun, and Chang}]{Hu2021Relation}
Ziniu Hu, Yizhou Sun, and Kai-Wei Chang. 2021.
\newblock \href {https://doi.org/10.18653/v1/2021.findings-emnlp.292}
  {Relation-guided pre-training for open-domain question answering}.
\newblock In \emph{Findings of the Association for Computational Linguistics:
  EMNLP 2021}, pages 3431--3448, Punta Cana, Dominican Republic. Association
  for Computational Linguistics.

\bibitem[{Huang et~al.(2018)Huang, Zhu, Shen, and Chen}]{huang2017fusionnet}
Hsin{-}Yuan Huang, Chenguang Zhu, Yelong Shen, and Weizhu Chen. 2018.
\newblock \href {https://openreview.net/forum?id=BJIgi\_eCZ} {Fusionnet: Fusing
  via fully-aware attention with application to machine comprehension}.
\newblock In \emph{6th International Conference on Learning Representations,
  {ICLR} 2018, Vancouver, BC, Canada, April 30 - May 3, 2018, Conference Track
  Proceedings}. OpenReview.net.

\bibitem[{Joshi et~al.(2017)Joshi, Choi, Weld, and
  Zettlemoyer}]{joshi2017triviaqa}
Mandar Joshi, Eunsol Choi, Daniel Weld, and Luke Zettlemoyer. 2017.
\newblock \href {https://doi.org/10.18653/v1/P17-1147} {{T}rivia{QA}: A large
  scale distantly supervised challenge dataset for reading comprehension}.
\newblock In \emph{Proceedings of the 55th Annual Meeting of the Association
  for Computational Linguistics (Volume 1: Long Papers)}, pages 1601--1611,
  Vancouver, Canada. Association for Computational Linguistics.

\bibitem[{Kwiatkowski et~al.(2019)Kwiatkowski, Palomaki, Redfield, Collins,
  Parikh, Alberti, Epstein, Polosukhin, Devlin, Lee
  et~al.}]{kwiatkowski2019natural}
Tom Kwiatkowski, Jennimaria Palomaki, Olivia Redfield, Michael Collins, Ankur
  Parikh, Chris Alberti, Danielle Epstein, Illia Polosukhin, Jacob Devlin,
  Kenton Lee, et~al. 2019.
\newblock Natural questions: A benchmark for question answering research.
\newblock \emph{Transactions of the Association for Computational Linguistics},
  7:453--466.

\bibitem[{Li et~al.(2020)Li, Wang, Dong, Wei, and Xu}]{Li2020Harvesting}
Zhongli Li, Wenhui Wang, Li~Dong, Furu Wei, and Ke~Xu. 2020.
\newblock \href {https://doi.org/10.18653/v1/2020.acl-main.600} {Harvesting and
  refining question-answer pairs for unsupervised {QA}}.
\newblock In \emph{Proceedings of the 58th Annual Meeting of the Association
  for Computational Linguistics}, pages 6719--6728, Online. Association for
  Computational Linguistics.

\bibitem[{Pan et~al.(2021)Pan, Hang, Qi, Shah, Potdar, and
  Yu}]{Pan2021Multilingual}
Lin Pan, Chung-Wei Hang, Haode Qi, Abhishek Shah, Saloni Potdar, and Mo~Yu.
  2021.
\newblock \href {https://doi.org/10.18653/v1/2021.naacl-main.20} {Multilingual
  {BERT} post-pretraining alignment}.
\newblock In \emph{Proceedings of the 2021 Conference of the North American
  Chapter of the Association for Computational Linguistics: Human Language
  Technologies}, pages 210--219. Association for Computational Linguistics.

\bibitem[{Raffel et~al.(2020)Raffel, Shazeer, Roberts, Lee, Narang, Matena,
  Zhou, Li, and Liu}]{Raffel2020T5}
Colin Raffel, Noam Shazeer, Adam Roberts, Katherine Lee, Sharan Narang, Michael
  Matena, Yanqi Zhou, Wei Li, and Peter~J. Liu. 2020.
\newblock Exploring the limits of transfer learning with a unified text-to-text
  transformer.
\newblock \emph{Journal of Machine Learning Research}, 21(140):1--67.

\bibitem[{Rajpurkar et~al.(2018)Rajpurkar, Jia, and Liang}]{rajpurkar2018know}
Pranav Rajpurkar, Robin Jia, and Percy Liang. 2018.
\newblock Know what you don’t know: Unanswerable questions for squad.
\newblock In \emph{Proceedings of the 56th Annual Meeting of the Association
  for Computational Linguistics (Volume 2: Short Papers)}, pages 784--789.

\bibitem[{Rajpurkar et~al.(2016)Rajpurkar, Zhang, Lopyrev, and
  Liang}]{rajpurkar2016squad}
Pranav Rajpurkar, Jian Zhang, Konstantin Lopyrev, and Percy Liang. 2016.
\newblock Squad: 100,000+ questions for machine comprehension of text.
\newblock In \emph{Proceedings of the 2016 Conference on Empirical Methods in
  Natural Language Processing}, pages 2383--2392.

\bibitem[{Ram et~al.(2021)Ram, Kirstain, Berant, Globerson, and
  Levy}]{ram2021few}
Ori Ram, Yuval Kirstain, Jonathan Berant, Amir Globerson, and Omer Levy. 2021.
\newblock \href {https://doi.org/10.18653/v1/2021.acl-long.239} {Few-shot
  question answering by pretraining span selection}.
\newblock In \emph{Proceedings of the 59th Annual Meeting of the Association
  for Computational Linguistics and the 11th International Joint Conference on
  Natural Language Processing (Volume 1: Long Papers)}, pages 3066--3079,
  Online. Association for Computational Linguistics.

\bibitem[{Trischler et~al.(2017)Trischler, Wang, Yuan, Harris, Sordoni,
  Bachman, and Suleman}]{trischler2017newsqa}
Adam Trischler, Tong Wang, Xingdi Yuan, Justin Harris, Alessandro Sordoni,
  Philip Bachman, and Kaheer Suleman. 2017.
\newblock Newsqa: A machine comprehension dataset.
\newblock \emph{ACL 2017}, page 191.

\bibitem[{Wu et~al.(2021)Wu, Ouyang, Ziegler, Stiennon, Lowe, Leike, and
  Christiano}]{wu2021recursively}
Jeff Wu, Long Ouyang, Daniel~M Ziegler, Nisan Stiennon, Ryan Lowe, Jan Leike,
  and Paul Christiano. 2021.
\newblock Recursively summarizing books with human feedback.
\newblock \emph{arXiv preprint arXiv:2109.10862}.

\bibitem[{Yang et~al.(2021)Yang, Wei, Jiao, Jiang, and Yang}]{Yang2021xmoco}
Nan Yang, Furu Wei, Binxing Jiao, Daxing Jiang, and Linjun Yang. 2021.
\newblock \href {https://doi.org/10.18653/v1/2021.acl-long.477} {x{M}o{C}o:
  Cross momentum contrastive learning for open-domain question answering}.
\newblock In \emph{Proceedings of the 59th Annual Meeting of the Association
  for Computational Linguistics and the 11th International Joint Conference on
  Natural Language Processing (Volume 1: Long Papers)}, pages 6120--6129.
  Association for Computational Linguistics.

\bibitem[{Yang et~al.(2018)Yang, Qi, Zhang, Bengio, Cohen, Salakhutdinov, and
  Manning}]{yang2018hotpotqa}
Zhilin Yang, Peng Qi, Saizheng Zhang, Yoshua Bengio, William Cohen, Ruslan
  Salakhutdinov, and Christopher~D Manning. 2018.
\newblock Hotpotqa: A dataset for diverse, explainable multi-hop question
  answering.
\newblock In \emph{Proceedings of the 2018 Conference on Empirical Methods in
  Natural Language Processing}, pages 2369--2380.

\bibitem[{Zhang et~al.(2022)Zhang, Zhu, Song, Koniusz, and
  King}]{DBLP:conf/kdd/ZhangZSKK22}
Yifei Zhang, Hao Zhu, Zixing Song, Piotr Koniusz, and Irwin King. 2022.
\newblock \href {https://doi.org/10.1145/3534678.3539425} {{COSTA:}
  covariance-preserving feature augmentation for graph contrastive learning}.
\newblock In \emph{{KDD} '22: The 28th {ACM} {SIGKDD} Conference on Knowledge
  Discovery and Data Mining, Washington, DC, USA, August 14 - 18, 2022}, pages
  2524--2534. {ACM}.

\end{thebibliography}

\appendix
\label{sec:appendix}
\section{Experimental Settings}\label{experimental_setting}
\subsection{Pre-Training Dataset}\label{pretraining_dataset}
\begin{description}[leftmargin=0pt,listparindent=\parindent,parsep=0pt]
\item[Cloze-like Samples.] We follow the method proposed by SSPT~\cite{glass2019span} to generate cloze-like pre-training instances. Specifically, we first choose a sentence in the Wikipedia corpus and randomly replace an entity or noun phrase with the special token [BLANK]. The new sentence will be considered as a query. Based on the query, we retrieve a paragraph that contains the masked term as its corresponding passage. Due to the time and computational resource limitations, we finally collect 5 million unique pre-training examples in English for our pre-training, which is one-sixth of the original SSPT method.
\item[Natural Samples.] For each $s^{\mathrm{cloze}} = (q_{\mathrm{c}}, p, a)$, we use a public T5-base model\footnote{\url{https://huggingface.co/valhalla/t5-base-qg-hl}} to generate natural queries $q_{\mathrm{n}}$ given $(p, a)$. It is fine-tuned for answer-aware question generation using the SQuADv1.1 train set. If answer $a$ is empty, $q_{\mathrm{n}}$ is not generated and loss functions related to $s^{\mathrm{natural}}$ will not be included during training. We do not apply extra filtering or other quality control operations for generated questions.
\end{description}

\subsection{Evaluation}
We use the official model evaluation scripts of SQuADv1.1\footnote{\url{https://github.com/allenai/bi-att-flow/blob/master/squad/evaluate-v1.1.py}} and  MRQA\footnote{\url{https://github.com/mrqa/MRQA-Shared-Task-2019}} to calculate the F1 and EM metrics between predictions and ground truth answers. Besides, we employ zero-shot F1 and zero-shot EM to quantify the performance of models without fine-tuning. Official training and development sets are used in all experiments.

\section{Implementation Details}\label{implementation_detail}
All model architectures are based on the 12-layer BERT-base model with 110M parameters, and we employed Google’s original implementation of the BERT model published in the Huggingface\footnote{\url{https://huggingface.co/docs/transformers/v4.21.2/en/model_doc/bert\#transformers.BertForQuestionAnswering}} to build up our encoder. The training batch size for pre-training and fine-tuning are 32 and 8. Total training steps for MCROSS and other baseline models are both kept to 156,250. The max sequence length of the transformer encoder is 384. In all experiments, we use the Adam optimizer with a learning rate of 2e-5. The MoCo momentum $m$ is set to 0.999, queue size 32,000, moco ratio $\lambda_{\mathrm{MoCo}}$ 1.0 and temperature $\tau$ 0.05. Models are implemented on PyTorch. When the passage length exceeds the max sequence length, we follow the sliding window strategy proposed for BERT transformers in both fast/slow encoder structures.

During QA prediction, we apply the constraint $st_a \leq en_a$ to filtering out implausible answer spans and then select the ones with the highest joint probability $p(st_a) \times p(en_a)$ as results.
\section{Design Choice: KL Divergence}\label{design_choice}
The performance gain of the MCROSS(UNI) method choosing KL divergence $\mathcal{D}_{\text{KL}}$ over cosine similarity  $\mathcal{D}_{\text{Cosine}}$ as similarity function $\mathcal{D}$ is shown in Table~\ref{d-compare-table}.

\begin{table}[!t]
\small
\centering
\setlength{\abovecaptionskip}{0.1cm}
\setlength{\belowcaptionskip}{-0.5cm}
\begin{tabular}{ccccc}
\hline
\bm{$\mathcal{D}$} & \textbf{Dataset} & \textbf{F1} & \textbf{EM}\T\B\\
\hline
\multirow{3}*{$\mathcal{D}_{\text{KL}}$} & SQuADv1.1 & \textbf{88.95} & \textbf{81.60}\T\B\\
~ & NewsQA & \textbf{66.45} & \textbf{51.33}\B\\
~ & TriviaQA & \textbf{72.06} & \textbf{66.41}\B\\
\hline
\multirow{3}*{$\mathcal{D}_\text{Cosine}$} & SQuADv1.1 & 88.47 & 80.79\T\B\\
~ & NewsQA & 65.98 & 50.40\B\\
~ & TriviaQA & 69.73 & 64.24\B\\
\hline
\end{tabular}
\caption{\label{d-compare-table}Comparison between choosing $\mathcal{D}_{\text{KL}}$ and $\mathcal{D}_\text{Cosine}$} 
\end{table}

\section{Space and Time Complexity}\label{space_time_complexity}
The additional parameters of the answer prediction layer come from Eq.~(\ref{eq:answer_span_predict}). The number of parameters is $(768+1) * 2=1,538$, which is negligible in front of the parameters in BERT-base~(110M). 
%SSPT(64), SSPT(64)$^\dag$, MCROSS(BI), and MCROSS(w/o $\mathcal{L}_{\mathrm{MoCo}}$)'s training time is at the same scale. 
Training of the SSPT baselines and the MCROSS method took approximately 40 hours each on 8 V-100 GPUs. %In addition, training of SSPT, SSPT$^\dag$, and MCROSS(UNI) costs an equivalent amount of time and takes around 32 hours on 8 V-100.

\end{document}